\documentclass{article}

\usepackage{arxiv}

\usepackage{multirow}
\usepackage{array}
\usepackage[ruled,vlined]{algorithm2e}
\usepackage{amsmath}
\usepackage{amssymb}

\newcommand{\fooAlter}{\hspace{-2pt}
\textcolor{blue}{$\bullet$} \hspace{5pt}}
\usepackage{xcolor}

\usepackage[utf8]{inputenc} 
\usepackage[T1]{fontenc}    
\usepackage{hyperref} 
\usepackage{scalerel,stackengine}
\stackMath
\newcommand\reallywidehat[1]{%
\savestack{\tmpbox}{\stretchto{%
  \scaleto{%
    \scalerel*[\widthof{\ensuremath{#1}}]{\kern-.6pt\bigwedge\kern-.6pt}%
    {\rule[-\textheight/2]{1ex}{\textheight}}
  }{\textheight}%
}{0.5ex}}%
\stackon[1pt]{#1}{\tmpbox}%
}

\usepackage{graphicx}
\usepackage[export]{adjustbox}
\usepackage{xcolor}
\usepackage{subfigure}
\usepackage{colortbl}
\usepackage{paralist}
\usepackage{bbm}
\usepackage[arabic,farsi,english]{babel}
\usepackage{url}            
\usepackage{booktabs}  
\usepackage[ruled,vlined]{algorithm2e}
\usepackage{amsfonts}       
\usepackage{nicefrac}       
\usepackage{microtype}      
\usepackage{lipsum}	
\usepackage{xcolor}
\title{Auditing and Robustifying COVID-19 Misinformation Datasets via Anticontent Sampling}

\author{
  Clay H. Yoo\\
  {Carnegie Mellon University}\\
  \texttt{hyungony@andrew.cmu.edu} \\
 \And
Ashiqur R. KhudaBukhsh\thanks{This work is accepted at AAAI 2023 (Robust and Safe AI track). Ashiqur R. KhudaBukhsh is the corresponding author.} \\
  {Rochester Institute of Technology} \\
  \texttt{axkvse@rit.edu}
} 

\begin{document}
\maketitle

\begin{abstract}

This paper makes two key contributions. First, it argues that highly specialized rare content classifiers trained on small data typically have limited exposure to the richness and topical diversity of the negative class (dubbed anticontent) as observed in the wild. As a result, these classifiers' strong performance observed on the test set may not translate into real-world settings. In the context of COVID-19 misinformation detection, we conduct an in-the-wild audit of multiple datasets and demonstrate that models trained with several prominently cited recent datasets are vulnerable to anticontent when evaluated in the wild. Second, we present a novel active learning pipeline that requires zero manual annotation and iteratively augments the training data with challenging anticontent, robustifying these classifiers.

\end{abstract}


\section{Introduction}

During the early phase of COVID-19 pandemic, the scientific community shared several COVID-19 misinformation datasets (see, e.g.,~\cite{Micallef2020TheRO,Memon2020CharacterizingCM,Alam2020FightingTC,Cheng2021ACR,Cui2020CoAIDCH}) to tackle the infodemic~\cite{cinelli2020covid} within a short turnaround time. While these datasets provided great value in jump-starting the battle against COVID-19 misinformation, follow-on behavioral science studies have started appearing that rely on models trained on these datasets~\cite{Verma2022ExaminingTI}. A precise understanding of these datasets' effectiveness is thus necessary to trust the broad societal conclusions that depend on the reliability of these models. 

\noindent\begin{figure*}[htb]
\centering
{\fcolorbox{red!80}{blue!25}{%
    \minipage[t]{\dimexpr0.96\linewidth-0\fboxsep-2\fboxrule\relax}
     \noindent\textcolor{black!70}{\textbf{Key idea:}} We leverage a simple assumption to conduct in-the-wild audit of COVID-19 misinformation datasets: \emph{before COVID-19, a social media post is highly unlikely to express topics related to COVID-19 misinformation.} We further tap into the richness and diversity of a vast amount of implicitly labeled (as negative, or non-misinformation) pool of social media posts (dubbed anticontent) and design an active learning framework that requires no human annotation. We start with an annotated COVID-19 misinformation dataset drawn from multiple well-known COVID-19 misinformation datasets. At each step, we train a model on labeled data and evaluate on the pool of implicitly labeled anticontent instances. Instances that are classified as positives (i.e. COVID-19 misinformation) with high confidence are added as challenging negatives to augment the train data.
    \endminipage}
}
\end{figure*}

\noindent\textbf{\textit{How do we conduct external audits of these misinformation datasets to estimate how well models trained on these datasets perform in the wild?}} In the responsible AI literature, internal audits of datasets presenting guidelines for  data dissemination~\cite{Gebru2018DatasheetsFD} and resource reproducibility practices~\cite{gebru2021datasheets} have positively contributed to robust AI efforts. For other rare class classification tasks such as hate speech detection, cross-dataset generalization has been studied before~\cite{arango2019hate}. In-the-wild robustness audits of datasets is an under-explored area. However, instances exist that rare-class classification methods may get completely blindsided when presented with adversarial examples~\cite{DBLP:conf/aaai/SarkarK21}.  

A limiting factor to conducting large-scale in-the-wild audit is the lack of ground truth. In this paper, we present a framework to conduct in-the-wild audit of COVID-19 misinformation datasets. Our framework does not require annotated examples. The heart of the framework lies in a simple yet powerful observation: \emph{before COVID-19, practically no COVID-19 misinformation can exist}. We show that classifiers trained on a broad range of heavily cited COVID-19 misinformation datasets often predict an alarming fraction of social media posts from the pre-COVID-19 era as COVID-19 misinformation, casting serious doubt on how useful they can be when deployed in the wild. Our experiments indicate that testing in the pre-COVID-19 era can present an elegant solution to estimating in-the-wild performance and evaluating real-world deployability. 

\subsection{Anticontent}
A realistic assumption in any social web platform with broad participation is that much of the user-generated content will possibly be \emph{non-misinformation}. Hence, even though discussions related to COVID-19 took the center stage during the pandemic, from the kneeling-down controversy of the NFL players to Ruth Bader Ginsburg's failing health; from Novak Djokovic's unfortunate expulsion
from the US open to the former president's misspelled Tweet, the diverse set of discussions that are \textbf{\emph{not misfinformation}}, is what we call anticontent. This anticontent is far too topically diverse to be effectively captured in a small labeled data scenario.

\noindent\textit{\textbf{What makes anticontent challenging?}} Anticontent can pose a challenge to in-the-wild COVID-19 misinformation classification in two possible ways. The first one is caused by shortcut learning~\cite{DBLP:journals/natmi/GeirhosJMZBBW20}. A classifier trained on a small dataset may pick up certain quirks of the dataset and incorrectly generalize. For instance, a model trained on a COVID-19 misinformation dataset where the keyword \texttt{hoax} is always associated with positives, will have a hard time when it encounters a linguistic black swan in the form of \emph{this impeachment is a hoax}. A classic example of such shortcut learning happened when hate speech classifiers mistook harmless chess discussions as racist because almost all examples in the train set had \texttt{black}, \texttt{white}, \texttt{kill}, \texttt{capture}, \texttt{threat} attached to racially charged social media posts~\cite{chess2021,DBLP:conf/aaai/SarkarK21}.  

The second situation may arise when the classifier encounters an out-of-domain example and has to spit out a prediction nonetheless. As we already mentioned, the richness and diversity of real-world social media discussions are hard to capture within small-sized misinformation datasets. Precisely due to its sharpened focus on exemplifying \emph{what is misinformation}, \emph{what is \textbf{not} misinformation} often remains under-specified. By letting the models choose their own Achilles heel through our iterative framework, we bolster our models against both types of anticontent.

\noindent\textit{\textbf{How can we leverage anticontent to robustify classifiers?}} Intuitively, an iterative learning framework akin to active learning~\cite{settles2009active} where challenging anticontent instances, guided by model prediction, are added to the training dataset, could be a viable path to robustify content classifiers. In this work, we propose an active learning framework that completely bypasses the annotator requirement of active learning through sampling from an implicitly labeled pool -- social media discussions from pre-COVID-19 time. To our knowledge, this is the first active learning framework that requires zero manual annotation.\footnote{As it will be evident later, our active learning framework \emph{does} need a small seed set of labeled examples to generate the initial model. However, the pipeline does not need any manual annotation in the subsequent learning steps. }

\subsection{Contributions}
Our contributions are the following: 

\noindent\fooAlter \textbf{\textit{Robustness Audit:}} We introduce a novel approach to auditing the in-the-wild performance of COVID-19 misinformation classifiers. In this approach, we evaluate these classifiers on a social media dataset predating COVID-19. A high false positive rate indicates vulnerability to anticontent and less suitability for deployment in the wild. 

\noindent\fooAlter \textbf{\textit{Robust AI Method:}} We present a novel active learning framework using implicit labels (\texttt{ALIL}) that requires zero human annotation since it samples from an implicitly labeled anticontent pool. Our method demonstrates substantial performance gain over classifiers trained solely on previously published COVID-19 misinformation datasets.   


\section{Active Learning Framework}

\paragraph{Background.} \textit{Active learning} is a powerful and well-established form of supervised machine learning technique~\cite{settles2009active}. It is characterized by the interaction between the learner, aka the classifier, and the teacher (oracle or labeler or annotator) during the learning process. \emph{Pool-based active learning} is a popular variant. In this setting, the learner is initially trained on a small seed set of labeled examples and has access to a large collection of unlabeled samples. At each iteration, the learner employs a sampling strategy to select an unlabeled sample and requests the supervisor to label it (in agreement with the target concept). The dataset is augmented with the newly acquired label, and the classifier is retrained on the augmented dataset. The sequential label-requesting and re-training process continues until some halting condition is reached (e.g., annotation budget is expended or the classifier has reached some target performance). At this point, the algorithm outputs a classifier, and the objective of this classifier is to closely approximate the (unknown) target concept in the future. The key goal of active learning is to reach a strong performance at the cost of fewer labels through the active participation of the learner. Since training and inference on a very large pool of samples can be computationally prohibitive, lines of work have examined the trade-offs of batch active learning~\cite{yang2013buy}. We follow the pool-based batch active learning pipeline where instead of requesting one label at a time, the learner requests labels in batches.   

\paragraph{Active Learning with Implicit Labels.}

Following traditional pool-based batch active learning, we start with a learner trained on a seed set of labeled examples. Let $\textit{batchSize}$ denote the configurable hyperparameter that specifies the number of samples added in each iteration. The large \emph{unlabeled} pool of samples the learner has access to consists of social media posts from pre-COVID-19 era. As per our assumption,  all the comments in this pool are implicitly labeled as negatives. We next sample the top $\textit{batchSize}$ samples with the highest predicted probability (as COVID-19 misinformation) computed by our learner. These samples are essentially highly challenging anticontent that can impact the classifier's performance in the wild. We add these samples to our training dataset as negatives, retrain our model on the augmented dataset, and loop through this cycle till a halting criterion is reached. Note that, in contrast with adversarial attacks (e.g.,~\cite{he2021petgen}), our approach adds real-world anticontent instances. Algorithm~\ref{algo:augment} presents a formal description. Hyperparameter choices are described in the experimental setup section.  

\begin{algorithm}[t]
\DontPrintSemicolon
\SetAlgoLined

\textbf{Input:} \\
$\mathcal{M}_0 := $ baseline classifier \\
$\mathcal{D}_\textit{train} :=$ train data\\
$\mathcal{D}_\textit{valid} :=$ validation data \\
$\mathcal{D}_\textit{implicit} :=$ implicitly labeled data \\

\textbf{Initialize:} \\
$prevScore  = -1$ \\
$bestValidationScore  = 0$ \\
$\mathcal{M}^* = \mathcal{M}_0$\\

\textbf{Procedure:} \\

\While{$bestValidationScore > prevScore$}
{
    $c = 0$ \\
    $\mathcal{D}_\textit{augment} = \emptyset$ \\
    $p =$ getPositiveProbability($\mathcal{M}^*$, $\mathcal{D}_\textit{implicit}$) \\
    $\mathbb{I}_m =$ argsort$(p,$ \text{ascending=False}$)$ \\
    \While{$c < batchSize$}
    {
        $\mathcal{D}_\textit{augment} = \mathcal{D}_\textit{augment} \cup \{\mathcal{D}_\textit{implicit}[\mathbb{I}_m[c]]\}$ \\
        $c \mathrel{+}= 1$ \\
    }
    $\mathcal{D} = \mathcal{D}_\textit{train} \cup \mathcal{D}_\textit{valid}\cup \mathcal{D}_\textit{augment} $ \\ 
    $\mathcal{D}_\textit{train}, \mathcal{D}_\textit{validation} =$ split$(\mathcal{D})$ \\ 
    $prevScore = bestValidationScore$ \\
    $\mathcal{M}_1, bestValidationScore =$ train$(\mathcal{D}_\textit{train}, \mathcal{D}_\textit{valid})$ \\
    \If{bestValidationScore $>$ prevScore}
    {
        $\mathcal{M}^* = \mathcal{M}_1$ \\
        $\mathcal{D}_\textit{implicit} = \mathcal{D}_\textit{implicit} \backslash \mathcal{D}_\textit{augment}$ \\
    }
}

\textbf{Output:} $\mathcal{M}^*$\\
\caption{\\{ {\small{$\textit{\texttt{ALIL}}$($\mathcal{M}_0$, $\mathcal{D}_\textit{train}$, $\mathcal{D}_\textit{valid}$, $\mathcal{D}_\textit{implicit}$, $batchSize$)}}}}

\label{algo:augment}

\end{algorithm}

\section{Related Work}
\label{sec:related-work}

Our work contains flavor of multiple well-established frameworks:  domain generalization~\cite{blanchard2011generalizing}; domain adaptation~\cite{blitzer2006domain}; and active learning~\cite{settles2009active}. Given the extensive literature in this field, we do not aim to be extensive and point to high-quality surveys whenever possible. 

Domain generalization, also known as out-of-distribution (OOD) generalization, is a widely studied machine learning research area~\cite{krueger2021out,teney2020unshuffling,hendrycks2021many}. \cite{shen2021towards} present a comprehensive survey. Similar to the extensive empirical evaluation conducted on a broad range of hate speech datasets~\cite{arango2019hate}, we conduct OOD generalization experiments (Secion~\ref{sec:OODGeneralization}) on COVID-19 misinformation datasets. Our evaluation framework on pre-COVID-19 content is also essentially geared towards estimating OOD generalization using implicit labels. However, to our knowledge, leveraging the temporal structure of an exogenous shock to construct a vast pool of implicitly labeled samples, and then evaluating OOD generalization, is a novel research direction.   

A key distinction between domain adaptation and domain generalization is that the former has access to samples in the target domain (labeled~\cite{DBLP:conf/acl/Daume07, plank2011domain} or unlabeled~\cite{ramponi2020neural}) unlike the latter. Our approach is thus more akin to domain adaptation. Again, what sets us apart is our novel use of pre-COVID-19 data with implicit labels. Following traditional self-training literature~\cite{hearst1991noun,zhou2012self}, we indeed select the unlabeled samples that are most confidently classified as positives. However, we flip their labels to negatives when we add them to our training set. 

We drew inspiration from several existing lines of active learning research for constructing our misinformation classifier~\cite{roy2001,baram2003,donmez2007,settles2009active}. Since sequentially labeling and retraining models may not be practically feasible, following~\cite{yang2013buy}, we adopted a batch active learning setting to expand our pool of labeled samples. Among the extensive literature of active learning query strategies~\cite{lewis1994sequential, zhu2009active, sindhwani2009uncertainty, attenberg2010unified}, we consider minority class certainty sampling~\cite{sindhwani2009uncertainty, attenberg2010unified} in our framework. Unlike more traditional applications of this sampling technique to address class imbalance~\cite{DBLP:conf/aaai/PalakodetyKCPKC20,DBLP:conf/ijcai/DuttaLNK22}, we use certainty sampling to detect challenging anticontent. 



\begin{table*}[htb]
\centering
\scriptsize
\begin{tabular}{l c c c c  c}
\toprule
& \multicolumn{5}{c}{Statistics}\\
Dataset & \# Citations & Source & Collection Period & \# Positives/\ \# Negatives/\ \# Total & Avg. Token Length    \\
\midrule

$\mathcal{D}_\textit{infodemic}$ & 23 & Twitter & Jan. -- May 2020 & 1,195 /\ 2,784 /\ 3,979 & $26.5_{13.4}$ \\

$\mathcal{D}_\textit{miscov19}$ & 107 & Twitter & Mar. -- Jun. 2020 & 615 /\ 2,853 /\ 3,468 & $28.7_{13.1}$ \\

$\mathcal{D}_\textit{rumor}$ & 22 & News Media \& Twitter & Jan. -- Apr. 2020 & 3,652 /\ 1,844 /\ 5,496 & $19.8_{13.6}$ \\

$\mathcal{D}_\textit{coaid}$ & 109 & News Media & Dec. 2019 -- Nov. 2020 & 1,723 /\ 590 /\ 2,313 & $11.2_{6.3}$ \\

$\mathcal{D}_\textit{disinfo}$ & 56 & Twitter & Jan. 2020 -- Mar. 2021 & 61 /\ 1,041 /\ 1,102 & $33.5_{10.7}$ \\

\midrule

Combined & $-$ & $-$ & $-$ & 6,113 /\ 10,245 /\ 16,358 & $23.0_{14.0}$ \\

\bottomrule
\end{tabular}
\caption{Statistics of datasets used for our research. Positives correspond to misinformation samples, while negatives to non-misinformation. The average and standard deviation (in subscript) of token length are measured by splitting preprocessed texts with a whitespace character. Citation counts are measured on 16 August 2022.}
\label{tab:dataset-stat}
\end{table*}



\section{Data}
\label{sec:data}

\paragraph{COVID-19 Misinformation Datasets.}\label{subsec:misinfo-data}

We audit five publicly available COVID-19 misinformation datasets. These datasets are: $\mathcal{D}_\textit{infodemic}$~\cite{Micallef2020TheRO}; $\mathcal{D}_\textit{miscov19}$~\cite{Memon2020CharacterizingCM}; $\mathcal{D}_\textit{rumor}$~\cite{Cheng2021ACR}; $\mathcal{D}_\textit{coaid}$~\cite{Cui2020CoAIDCH}; and $\mathcal{D}_\textit{disinfo}$~\cite{Alam2020FightingTC}. Table~\ref{tab:dataset-stat} presents the overall statistics. We collapse fine-grained labels into two broad categories: \emph{misinformation} and \emph{non-misinformation} and use standard preprocessing steps.  We denote train and validation sets as $\mathcal{D}^\textit{train}$ and test set as $\mathcal{D}^\textit{test}$. Moreover, a model trained with dataset $\mathcal{D}$ is denoted as $\mathcal{M}(\mathcal{D})$. 

\renewcommand{\tabcolsep}{1mm}
\begin{table*}[htb]
\centering
\small
\begin{tabular}{l c c c c c c c}
\toprule
& \multicolumn{5}{c}{Test dataset} & \multicolumn{2}{c}{Misinformation Rate}\\
Model & $\mathcal{D}_\textit{infodemic}$ & $\mathcal{D}_\textit{miscov19}$ & $\mathcal{D}_\textit{rumor}$ & $\mathcal{D}_\textit{coaid}$ & $\mathcal{D}_\textit{disinfo}$ & $\mathcal{D}_\textit{pre}^\textit{1m}$ & $\mathcal{D}_\textit{post}^\textit{1m}$ \\

\midrule

$\mathcal{M}(\mathcal{D}_\textit{infodemic}^\textit{train})$ & $89.7_{0.5}/78.7_{0.9}$ & $\mathbf{86.2_{0.6}}/23.4_{1.2}$ & $\mathbf{55.1_{0.6}}/39.7_{2.9}$ & $83.0_{1.1}/31.9_{0.9}$ & $\mathbf{94.2_{0.7}}/4.6_{1.7}$ & $1.7_{1.5}$ & $2.0_{1.4}$ \\

$\mathcal{M}(\mathcal{D}_\textit{miscov19}^\textit{train})$ & $\mathbf{81.3_{0.4}}/30.8_{5.0}$ & $93.9_{0.6}/69.4_{2.8}$ & $51.0_{0.3}/19.3_{3.0}$ & $\mathbf{83.7_{1.0}}/17.7_{1.3}$ & $91.8_{2.6}/5.4_{2.3}$ & $7.6_{2.8}$ & $6.9_{2.4}$\\

$\mathcal{M}(\mathcal{D}_\textit{rumor}^\textit{train})$ & $76.8_{2.0}/\mathbf{52.7_{1.1}}$ & $79.3_{2.6}/25.2_{1.1}$ & $77.5_{1.9}/89.9_{1.3}$ & $59.0_{4.9}/\mathbf{45.1_{1.4}}$ & $88.3_{1.8}/\mathbf{12.2_{2.6}}$ & $18.3_{4.6}$ & $18.8_{4.8}$ \\

$\mathcal{M}(\mathcal{D}_\textit{coaid}^\textit{train})$ & $48.0_{6.1}/47.3_{1.0}$ & $52.5_{7.0}/\mathbf{29.6_{0.9}}$ & $38.7_{2.7}/\mathbf{76.5_{0.4}}$ & $97.0_{0.6}/91.2_{1.8}$ & $44.9_{6.0}/11.5_{0.5}$ & $62.2_{6.1}$ & $60.5_{6.4}$ \\




\bottomrule
\end{tabular}
\caption{Mean and standard deviation (in subscript) of F-1 scores of two labels (non-/misinformation) trained and evaluated on each dataset over five seeds. When the train and test datasets are the same (diagonal entries), performance on the test set is reported. Otherwise, performance on the entire dataset is reported (non-diagonal entries). The best out-of-domain F-1 score of each label is boldfaced. We also report misinformation rates on 1 million randomly selected pre-COVID-19 comments ($\mathcal{D}_\textit{pre}^\textit{1m}$) and post-COVID-19 comments ($\mathcal{D}_\textit{post}^\textit{1m}$).
}
\label{tab:data-against-data-performance}
\end{table*}

\paragraph{YouTube Video Comments.}
\label{subsec:youtube-data}

For our in-the-wild audit, we consider a  dataset of YouTube comments from the official channels of four prominent US cable news networks: CNN, Fox News, MSNBC, and One American News Network (OANN)\footnote{Compared to the other three networks, OANN is a fringe network often harboring outlandish views on science and democracy. Through using OANN, we are in no way legitimizing OANN as a mainstream news network. We include OANN due to its well-documented history of propagating COVID-19 misinformation.  }~\cite{KhudaBukhsh2021FringeNN,KhudaBukhsh2021WeDS}. We consider this dataset for the following three reasons: broad participation, topical diversity, and substantial presence of discussions relevant to COVID-19.  


Overall, the evaluation dataset consists of 33.3 million comments on news videos starting from 25 January 2014 to 2 November 2020. We focus on two non-overlapping partitions of the dataset -- pre-COVID-19 ($\mathcal{D}_{\textit{pre}}$) and post-COVID-19 ($\mathcal{D}_{\textit{post}}$) -- separating before and after the outbreak of COVID-19. Following the Centers for Disease Control and Prevention (CDC) timeline\footnote{\url{https://www.cdc.gov/museum/timeline/covid19.html}}, we mark 12 December 2019 as the start date of COVID-19. To summarise, $\mathcal{D}_{\textit{pre}}$ spans between 25 January 2014 and 11 December 2019, while $\mathcal{D}_{\textit{post}}$ spans between 12 December 2019 and 2 November 2020.   

\noindent\textit{\textbf{How rare is COVID-19 misinformation in the wild?}} We randomly sample 5,000 comments from $\mathcal{D}_\textit{post}$ and annotate them as one of misinformation, non-misinformation, and unverifiable. We confirm the veracity of comments from the official websites of credible sources such as CDC, WHO, IFCN’s CoronaVirusFacts / DatosCorona-Virus alliance database, Reuters, WebMD and PolitiFact. 

Out of 5,000 comments, 4,904 are marked as non-misinformation, 83 as misinformation, and 13 as indeterminate. We discard 13 indeterminate comments and construct a test set of 4,904 comments ($\mathcal{D}_\textit{youtube}^\textit{test}$), yielding $1.7\%$ misinformation rate. Two annotators independently annotated with a post-annotation adjudication step to dissolve the disagreements with $\kappa = 0.82$~\cite{cohen1960coefficient}. 

These annotated 4,987 samples also serve another critical purpose. We now have a dataset that is markedly different both in terms of the relative proportion of misinformation and non-misinformation and (possibly) richness in anticontent, allowing us to estimate OOD performance.

\paragraph{Combining Datasets.} Seeking to build a robust classifier capturing a wide range of topics in COVID-19 misinformation, we combine five misinformation datasets into a single set. We first split each dataset using 80/10/10 (train/validation/test) ratio. We then concatenate five individual test sets and $\mathcal{D}_\textit{youtube}^\textit{test}$ to obtain 5,634 non-misinformation and 558 misinformation samples as a combined test set ($\mathcal{D}_\textit{combined}^\textit{test}$). Since our goal is to test the OOD performance, $\mathcal{D}_\textit{combined}^\textit{train}$ does not contain any of the labeled YouTube comments.

\renewcommand{\tabcolsep}{2mm}
\begin{table*}[htb]
\small
\centering
\begin{tabular}{|p{0.45\textwidth}|p{0.45\textwidth}|}
\hline
\multicolumn{1}{|c|}{$\mathcal{M}(\mathcal{D}_\textit{combined}^\textit{train})$} 
& \multicolumn{1}{c|}{$\mathcal{M}(\mathcal{D}_\textit{combined}^\textit{train} \cup \mathcal{D}_\textit{500}^\textit{ALIL})$} \\

\hline

\textit{at the ohio state university they limited domestic students and started to give those seats to the chinese}  
&  \textit{there is now ample evidence that the united states of america government through the national institute of health have committed an act of bio terror by funding an experiment in 2015 at the university of north carolina chapel hill that created sars covid 2 covid 19 \ldots}  \\

\hline

 \textit{obama was fully aware of covid 19 and also enabled the funding and creation aswell as selling it to china} 
& \textit{the plan is to infect as many americans with corona virus to flood the streets with regeneron} \\

\hline 

 \textit{joe biden and hussein obama directly funded the missiles that were launched at our servicemen with the pallets of cash they sent to iran} 
& \textit{no one is safe from this bioweapon virus it now has over 40 mutations young and old r now in great danger human depopulation is happening at a rapid pace hopefully there will be a vaccine created soon to eradicate it} \\

\hline

 \textit{john bolton said there are interest groups who profit on wars bombs and bullets etc are a source of money on wartime}  
&  \textit{would not be surprised if we hear the dems went to china and came back with the china bioweapon virus attack on purpoae} \\

\hline

 \textit{ronald reagan said if facism ever comes to the us it ll come in the form of liberalism}  
&  \textit{this is propaganda they eat those types of animals for years this was predicted by billy meier back in 1995 and he says quote a lung disease will also break out in humans through the guilt of china where bioweapons are being researched and a carelessness is releasing pathogens}  \\

\hline
\end{tabular}
\caption{Top five misinformation with the highest misinformation probability selected by the baseline model $(\mathcal{M}(\mathcal{D}_\textit{combined}^\textit{train}))$ versus the top comments selected by $\mathcal{M}(\mathcal{D}_\textit{combined}^\textit{train} \cup \mathcal{D}_\textit{500}^\textit{ALIL})$.}
\label{tab:top10}
\end{table*}

\section{Experimental Setup}\label{sec:ALILHyp}

We fine-tune \texttt{BERT}~\cite{Devlin2019BERTPO} when training all classifiers.

\paragraph{\texttt{ALIL} Setup.} For augmentation, we construct a dataset of 1 million randomly sampled YouTube comments from pre-COVID-19 era ($\mathcal{D}_\textit{pre}^\textit{1m}$) such that none of the comments in $\mathcal{D}_\textit{youtube}^\textit{test}$ is included. $\mathcal{D}_\textit{pre}^\textit{1m}$ is used as $\mathcal{D}_\textit{implicit}$ in Algorithm~\ref{algo:augment}. We set $\textit{batchSize} = 500$ ($\approx 4\%$ of the combined train data).
As a stopping criterion, we stop training if we observe a decrease in the validation F-1 score. 

\paragraph{Self-training Setup.} Since we do not assume there exists a temporal structure in YouTube comments, we randomly sample 1 million comments from $\mathcal{D}_\textit{post}$ such that it does not overlap with any comment in $\mathcal{D}_\textit{post}^{1m}$ or $\mathcal{D}_\textit{youtube}^\textit{test}$ and use it as an unlabeled pool. At each self-training iteration, we add 250 samples with the lowest misinformation probability as non-misinformation and 250 with the highest probability as misinformation to match the size of added samples same as \texttt{ALIL}. We avoid using $\mathcal{D}_\textit{pre}$ as a pool since adding 250 pre-COVID-19 comments as misinformation will likely harm the model performance.

\paragraph{Baselines.} We compare our \texttt{ALIL} against (1) a model trained with self-training and (2) a model trained on the seed set ($\mathcal{D}_\textit{combined}^\textit{train}$) without any data augmentation.

\section{Results}
\label{sec:result}

\subsection{In-the-wild Audit and OOD Generalization}
\label{sec:OODGeneralization}

We first examine OOD generalization through training models on a single dataset and testing on another. We could not obtain  $\mathcal{M}(\mathcal{D}_\textit{disinfo})$ since training didn't converge, possibly due to the scarcity of positive samples (5.5\%) obtained during rehydration. Table~\ref{tab:data-against-data-performance} summarizes the OOD generalization of the four datasets. We observe the in-domain performances are significantly better than the OOD performances.

Table~\ref{tab:data-against-data-performance} conducts an in-the-wild audit through evaluating on 1 million randomly sampled YouTube comments from pre-COVID-19 ($\mathcal{D}_\textit{pre}^\textit{1m}$) and post-COVID-19 ($\mathcal{D}_\textit{post}^\textit{1m}$) eras, respectively. We expect robust models to have (1) close to zero pre-COVID-19 misinformation rate; and (2) close to $1.7\%$ post-COVID-19 misinformation rate (estimated through manual inspection). We observe that two models trained on these datasets exhibit a pre-COVID-19 misinformation rate higher than 10\%, indicating susceptibility to anticontent and making them unsuitable for real-world deployment.



\renewcommand{\tabcolsep}{3mm}
\begin{table*}[htb]
\centering
\small
\begin{tabular}{l c c c c c}
\toprule
& \multicolumn{3}{c}{Performance on $\mathcal{D}^\textit{test}_\textit{combined}$} & \multicolumn{2}{c}{Misinformation Rate} \\
Model & Precision & Recall & F-1 & $\mathcal{D}^{1m}_\textit{pre}$ & $\mathcal{D}^{1m}_\textit{post}$ \\
\midrule

$\mathcal{M}(\mathcal{D}_\textit{combined}^\textit{train})$ & $91.3_{0.3}$ & $85.7_{2.6}$ & $87.7_{1.8}$ & $14.3_{3.4}$ & $13.6_{3.1}$ \\ 

\midrule

$\mathcal{M}(\mathcal{D}_\textit{combined}^\textit{train} \cup \mathcal{D}^\textit{baseline}_{500}$) & $94.1_{0.2}$ & $93.9_{0.5}$ & $94.0_{0.4}$ & $1.7_{0.5}$ & $3.4_{0.6}$ \\
$\mathcal{M}(\mathcal{D}_\textit{combined}^\textit{train} \cup \mathcal{D}^\textit{baseline}_{1000}$)  & $94.9_{0.3}$ & $95.0_{0.4}$ & $94.9_{0.4}$ & $0.7_{0.4}$ & $1.9_{0.7}$ \\

\midrule
$\mathcal{M}(\mathcal{D}_\textit{combined}^\textit{train} \cup \mathcal{D}^\textit{ALIL}_{500})$ & $94.1_{0.2}$ & $93.9_{0.5}$ & $94.0_{0.4}$ & $1.7_{0.5}$ & $3.4_{0.6}$ \\

$\mathcal{M}(\mathcal{D}_\textit{combined}^\textit{train} \cup \mathcal{D}^\textit{ALIL}_{1000})$ & \textbf{$95.2_{0.2}$} & \textbf{$95.3_{0.2}$} & \textbf{$95.2_{0.2}$} & \textbf{$0.2_{0.1}$} & \textbf{$1.4_{0.4}$} \\

\midrule

$\mathcal{M}(\mathcal{D}_\textit{combined}^\textit{train} \cup \mathcal{D}^\textit{random,pre}_{1000})$  & $93.9_{0.3}$ & $92.9_{1.1}$ & $93.3_{0.8}$ & $3.7_{1.2}$ & $5.0_{1.6}$ \\
$\mathcal{M}(\mathcal{D}_\textit{combined}^\textit{train} \cup \mathcal{D}^\textit{random,post}_{1000})$ & $94.0_{0.2}$ & $93.5_{0.4}$ & $93.7_{0.3}$ & $3.4_{0.4}$ & $4.0_{0.4}$ \\

\midrule

$\mathcal{M}(\mathcal{D}_\textit{combined}^\textit{train} \cup \mathcal{D}^\textit{self-train}_{500})$ & $91.1_{0.2}$ & $84.3_{0.7}$ & $86.7_{0.5}$ & $16.7_{0.0}$ & $16.3_{0.0}$  \\
$\mathcal{M}(\mathcal{D}_\textit{combined}^\textit{train} \cup \mathcal{D}^\textit{self-train}_{1000})$ & $90.8_{0.1}$ & $81.6_{1.1}$ & $84.8_{0.8}$ & $20.8_{0.0}$ & $19.7_{0.0}$ \\

\bottomrule
\end{tabular}
\caption{
Mean and standard deviation (in subscript) of precision, recall, and F-1 scores of two labels (non-/misinformation) evaluated on $\mathcal{D}_\textit{combined}^\textit{test}$ over five seeds. We also report misinformation rates on fixed sets of 1 million randomly selected pre-COVID-19 comments ($\mathcal{D}^{1m}_\textit{pre}$) and post-COVID-19 comments ($\mathcal{D}_\textit{post}^\textit{1m}$).
}
\label{tab:augment-model-performance}
\end{table*}

\subsection{Evaluating Active Learning with Implicit Labels}

\paragraph{Qualitative Evaluations.}  Table~\ref{tab:top10} shows the top five misinformation comments from $\mathcal{D}_\textit{post}^\textit{1m}$ assigned the highest probability by two different models. The left column contains predictions by a model trained on $\mathcal{D}_\textit{combined}^\textit{train}$ (baseline) without any augmentation of anticontent. The right column contains predictions by $\mathcal{M}(\mathcal{D}_\textit{combined}^\textit{train} \cup \mathcal{D}^{ALIL}_{500})$ obtained after the first iteration of \texttt{ALIL}. On the left side, we notice that one comment is both COVID-19-related and COVID-19 misinformation, while the other comments are unrelated to COVID-19. On the right side, however, we notice that the model can sift through anticontent far more effectively, detecting several misinformation posts.  

\begin{figure*}[htb]
    \centering
    \includegraphics[frame,height=6.5cm]{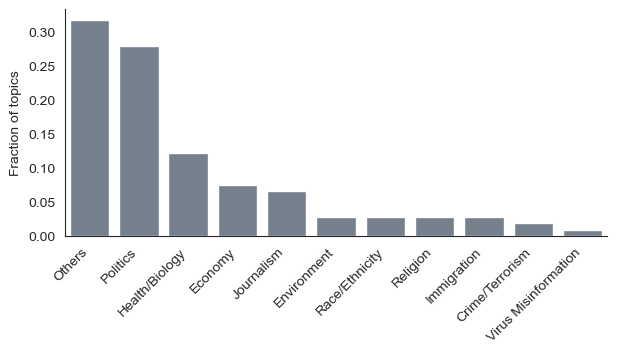}
    \caption{Distribution of annotated topics present in 200 randomly selected YouTube comments from $\mathcal{D}_{1000}^\textit{ALIL}$.}
    \label{fig:topic_distribution}
    \vspace{-0.1cm}
\end{figure*}

\paragraph{Quantitative Evaluations.} In order to demonstrate that iterative augmentation of the train data is more effective than augmenting with a larger \emph{batchSize} of samples added all at once, we construct two datasets with increasing number of $N \in \{500, 1000\}$ implicitly labeled pre-COVID-19 comments. $\mathcal{D}^\textit{baseline}_N$ is constructed by selecting pre-COVID-19 comments assigned the highest misinformation probabilities computed by the baseline model, i.e., $\mathcal{M}(\mathcal{D}_\textit{combined}^\textit{train})$, while $\mathcal{D}^\textit{ALIL}_N$ is constructed with \texttt{ALIL} (Algorithm~\ref{algo:augment}).
Note that $\mathcal{D}^\textit{baseline}_{500} \subset \mathcal{D}^{baseline}_\textit{1000}$, $\mathcal{D}^\textit{ALIL}_{500} \subset \mathcal{D}^\textit{ALIL}_{1000}$ and $\mathcal{D}^\textit{baseline}_{500} = \mathcal{D}^\textit{ALIL}_{500}$.

Table~\ref{tab:augment-model-performance} summarizes the performance of \texttt{ALIL}. We observe that the initial addition of 500 pre-COVID-19 comments achieves a substantial gain in performance, improving a weighted F-1 score from 87.7\% to 94.0\%, when evaluated on $\mathcal{D}_\textit{combined}^\textit{test}$. Pre-COVID-19 misinformation rate drastically decreases from 14.3\% to 1.7\% as well as the post-COVID-19 misinformation rate from 13.6\% to 3.4\%. In the second step of augmentation, models trained with $\mathcal{D}^\textit{ALIL}_{1000}$ achieves a better performance (95.2\% v.s. 94.9\%) and significantly lower pre-COVID-19 misinformation rate (0.2\% v.s. 0.7\%)  compared to the models trained with $\mathcal{D}^\textit{baseline}_{1000}$. We stop augmenting at 1,000 pre-COVID-19 comments since the average validation performance drops with $\mathcal{D}^\textit{ALIL}_{1500}$. With the addition of 1,000 samples, we observe that the post-COVID-19 misinformation rate gets closer to the approximated population misinformation rate of $1.7\%$. 

To test the effect of \textit{batchSize}, we ran the same experiment with $batchSize = 100$. \texttt{ALIL} halted at the fifth iteration, and it showed a similar pattern as reported in Table~\ref{tab:augment-model-performance}; improvement in performance and reduction in pre/post-COVID-19 misinformation rate.

We denote $\mathcal{D}_{N}^\textit{self-train}, N \in \{500, 1000\}$, as the augmentation sets found by self-training. We observe that performance rather decreases, and the misinformation rates for both pre-COVID-19 and post-COVID-19 eras increase. We tried (1) decreasing the batch size from 500 to 250, i.e., add 125 positives and 125 negatives at each iteration, and (2) changing the distribution to match that of $\mathcal{D}_\textit{combined}$, i.e., adding 190 positives and 310 negatives at each iteration, but the story stayed the same. 

\section{Conclusions and Discussions}
\label{sec:discussion}

While dataset audit is still in its infancy mainly focusing on fairness, transparency, and accountability~\cite{Gebru2018DatasheetsFD}, we highlight a relatively under-explored area of auditing in-the-wild reliability of COVID-19 misinformation datasets. As the field of information mining is moving towards dense inter-dependence, where a dataset proposed by one research group facilitates social investigations by other groups, testing the in-the-wild effectiveness of datasets will become increasingly important. We believe our work will open the gates for further innovative research in reliability audits for datasets.  

We also raise an important point that given misinformation is (hopefully) rare, and to operate successfully in a real-world setting with high topical diversity, a COVID-19 misinformation content classifier needs to both understand what \textbf{is} and \textbf{is not} misinformation. While a dataset of a few 1,000 samples can do a decent job in describing what is misinformation, the complement of the set, what we dub anticontent, has a richness extremely challenging to be captured well within a small dataset scenario. Arguing that before COVID-19, no social media content can express COVID-19 misinformation, we propose that testing on a large-scale dataset of pre-COVID-19 era social media discussions can shed light on content classifiers' vulnerability to anticontent. Based on this insight, we further propose an active learning framework that remarkably improves the content classifiers' performance in handling anticontent. And this performance boost comes with zero additional manual annotation.  

Our study raises the following points to consider.

\paragraph{Topical Diversity in Anticontent.} We annotate 200 random samples from $\mathcal{D}^\textit{ALIL}_{1000}$ into  one of the 10 topics. 
We do not intend to be formal or exhaustive, but rather to be illustrative of the broad range of topics present in anticontents that are misclassified as misinformation. Figure~\ref{fig:topic_distribution} shows that while politics is the dominating topic, the selected anticontent possesses a healthy diversity of a broad range of topics.




\paragraph{Misinformation about Previous Outbreaks.} Although there cannot exist COVID-19-specific misinformation in the pre-COVID-19 era, conspiracies, anti-vaccine sentiment and fake cure/treatment, topics that appear frequently in our combined dataset, existed well before COVID-19 from the previous outbreaks (e.g. SARS, Zika Virus, MERS). In fact, Figure~\ref{fig:topic_distribution} shows that two comments (1\%) are marked as misinformation related to non-COVID-19 diseases/viruses, where one comment discusses biohazard weapons related to tuberculosis and the other discusses how flu shot is related to Bill Gates as a medium of population control. Both topics appear as COVID-19 misinformation in our combined dataset, and adding these samples as non-misinformation to the train set for COVID-19 misinformation detection can either i) improve the classifier's performance by letting it learn how to distinguish between COVID-19 and other diseases (tuberculosis and flu in this case) or ii) confuse the classifier by providing similar comments with opposing labels. Although the impact of adding these samples may be small due to their size (1\%), figuring out precisely how they impact performance merits a deeper future exploration.

\paragraph{Generalizability to Tasks without Temporal Structure.} COVID-19 is an exogenous shock with a crisp temporal structure, making it straightforward to construct a set of anticontent predating COVID-19. \textbf{\textit{How can we use a similar approach to improve classifier performance in tasks without crisp temporal structures, such as detecting anti-Semitism, misogyny, or racism?}} Defining implicitly labeled anticontent for such problems is not as straightforward as COVID-19 misinformation detection since there are no well-established points of the time these phenomena started. 

We present a path forward in incorporating our approach to combat these types of inappropriate content. For instance, a strictly moderated community, $\mathcal{C}$, designated as a safe space for say, the LGBTQ+ community, is highly unlikely to have a large number of anti-LGBTQ+ posts. A hate speech classifier designed to protect the LGBTQ+ community can adopt \texttt{ALIL}. Even if hate speech exists in $\mathcal{C}$, augmenting the train set may still lead to performance gain so long as the overall proportion of hate speech is low.

To test our hypothesis, we assume that there exists no temporal structure in COVID-19 and that the overall misinformation rate is low, possibly close to 1.7\% as computed from $\mathcal{D}_\textit{youtube}^\textit{test}$. To augment $\mathcal{D}_\textit{combined}^\textit{train}$, we randomly sample 1,000 post-COVID-19 comments and implicitly label them as non-misinformation ($\mathcal{D}^\textit{random,post}_{1000}$). Even though there may exist misinformation among 1,000 samples, we observe that augmentation leads to (1) an improvement in weighted F-1 score from 87.7\% to 93.7\%, (2) a reduction of pre-COVID-19 misinformation rate from 14.3\% to 3.4\%, and (3) bringing post-COVID-19 misinformation rate closer to 1.7\% from 13.6\% to 4.0\% (Table~\ref{tab:augment-model-performance}). Although not as effective as using the temporal structure and \texttt{ALIL}, this shows that our approach of finding implicitly labeled anticontent in a low probability (of a positive label) setting can be extended to solving more general problems than a special case like detecting COVID-19 misinformation.

\paragraph{Value in the Active Learning Framework.} Since the entire comment set predating COVID-19 could serve as anticontent, it is possible to augment the train data using randomly sampled pre-COVID-19 comments. One may then wonder if the active learning pipeline is adding any value to performance gain. We construct $\mathcal{D}^\textit{random,pre}_{1000}$ by randomly sampling 1,000 pre-COVID-19 comments from $\mathcal{D}_\textit{pre}$, ensuring that there is no overlap with comments in $\mathcal{D}_\textit{youtube}^\textit{test}$ and $\mathcal{D}_\textit{pre}^{1m}$. We observe that adding 1,000 random samples has similar performance effects in (1) improving F-1 score on $\mathcal{D}_\textit{combined}^\textit{test}$ from 87.7\% to 93.3\%, (2) reducing pre-COVID-19 misinformation rate from 14.3\% to 3.7\%, and (3) bringing the post-COVID-19 misinformation rate closer to 1.7\% from 5.0\% (Table~\ref{tab:augment-model-performance}). However, the improvement is less pronounced than that of using \texttt{ALIL}.

\paragraph{Rehydration of Tweets and its Potential Impact.} Our experiments are conducted on the currently available snapshots of the datasets. To preserve a user's \emph{right to be forgotten}, most of these datasets only provide Tweet IDs that need to be re-hydrated. Some of the samples present in the released datasets are no longer available due to reasons like deletion of posts (e.g. Twitter's COVID-19 misleading information policy) and user suspension. 



\section{Ethics Statement}

Although our focus is to build a robust COVID-19 misinformation classifier, any content filter can be inverted for malicious purposes. For example, a real-world application can utilize our misinformation classifier to effectively detect misinformation but maliciously choose to filter out non-misinformation and filter in misinformation. 




\section{Acknowledgements}
We thank Anirban Chowdhury, Rupak Sarkar, and Ritam Dutta for their input.



\end{document}